\documentclass[journal]{IEEEtran}
\ifCLASSINFOpdf
\else
\fi

\usepackage[numbers,sort]{natbib}
\usepackage{caption}
\usepackage{subcaption}
\usepackage{algorithmic}
\usepackage{times}
\usepackage{epsfig}
\usepackage{graphicx}
\usepackage{amsmath}
\usepackage[amsmath, thmmarks]{ntheorem}
\usepackage{amssymb}

\usepackage{xcolor}
\usepackage{soul}
\usepackage{float}
\usepackage{fixltx2e}

\newcommand{\ie}{\textit{i}.\textit{e}.}
\usepackage[breaklinks=true,bookmarks=false]{hyperref}
\begin{document}
%
\title{ A Deep Learning Algorithm for One-step Contour Aware Nuclei Segmentation of Histopathological Images}
%
%
%

\author{Yuxin Cui\textsuperscript{*},
Guiying Zhang\textsuperscript{*},
Zhonghao Liu,
Zheng Xiong,
Jianjun Hu\textsuperscript{\#},~\IEEEmembership{Member, ~IEEE}

\IEEEauthorblockA{Department of Computer Science and Engineering \\
University of South Carolina\\
Columbia, SC 29208\\
Email: \{yxcui,jianjunh\}@cse.sc.edu}

\thanks{* Equal contribution. \#Corresponding author.}

        }

\maketitle

\begin{abstract}
This paper addresses the task of nuclei segmentation in high-resolution histopathological images. We propose an automatic end-to-end deep neural network algorithm for segmentation of individual nuclei. A nucleus-boundary model is introduced to predict nuclei and their boundaries simultaneously using a fully convolutional neural network. Given a color normalized image, the model directly outputs an estimated nuclei map and a boundary map. A simple, fast and parameter-free post-processing procedure is performed on the estimated nuclei map to produce the final segmented nuclei. An overlapped patch extraction and assembling method is also designed for seamless prediction of nuclei in large whole-slide images. We also show the effectiveness of data augmentation methods for nuclei segmentation task. Our experiments showed our method outperforms prior state-of-the-art methods. Moreover, it is efficient that one 1000X1000 image can be segmented in less than 5 seconds. This makes it possible to precisely segment the whole-slide image in acceptable time. 
\end{abstract}


%
\IEEEpeerreviewmaketitle

\section{Introduction}
%
%
%
%
\IEEEPARstart{W}{ith} the progress of image processing and pattern recognition techniques, computer-assisted diagnosis(CAD) has been widely utilized to assist medical professionals to interpret medical images. Digital pathology,as an important aspect of CAD application, is earning more and more attention from both image analysis researchers and pathologists due to the advent of whole-slide imaging. Its aim is to acquire, manage and interpret pathology information generated from digitized glass slides, among which the development of computational algorithms to automatically analyze digital tissue images is the key. The potential applications of digital pathology span a wide range such as segmentation of desired regions or objects, counting normal or cancel cells, recognizing tissue structures, classifying cancer grades, prognosis of cancers, etc \cite{pmid24759275,pmid28066683}. It is able to dramatically decrease human's workload and has the potential to work better than pathologists due to its objectiveness in the interpretation. 

\begin{figure}[hbp]
  \begin{center}
    \begin{subfigure}[t]{0.20\textwidth}
      \includegraphics[width=3.5cm]{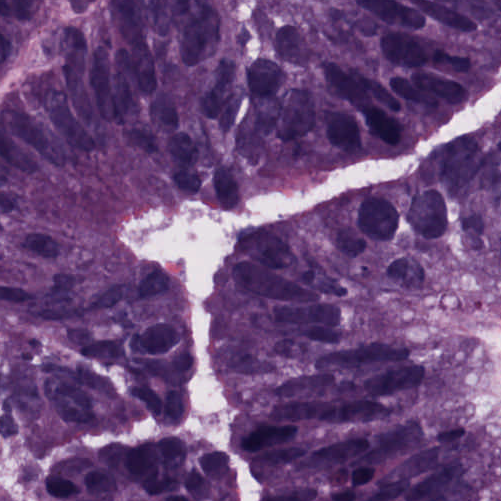}
      
      \caption{}
      \label{colon}
    \end{subfigure}
    \begin{subfigure}[t]{0.20\textwidth}
      \includegraphics[width=3.5cm]{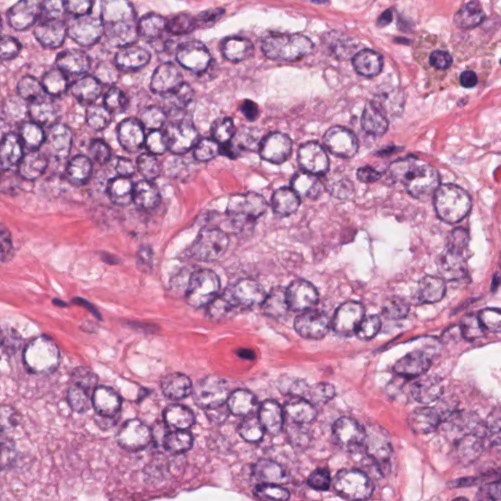}
      
      \caption{}
      \label{prostate}
    \end{subfigure}
    \begin{subfigure}[t]{0.20\textwidth}
      \includegraphics[width=3.5cm]{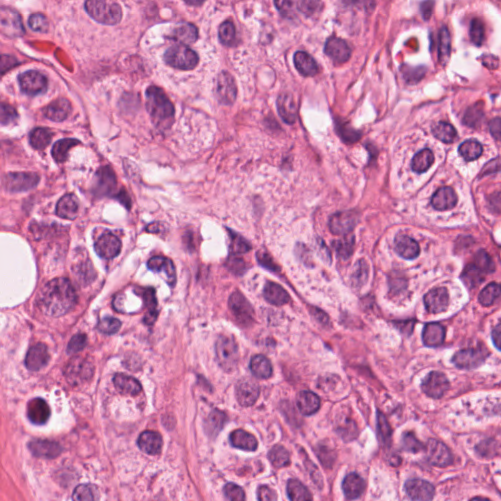}
      
      \caption{}
      \label{breast_cancer1}
    \end{subfigure}
    \begin{subfigure}[t]{0.20\textwidth}
      \includegraphics[width=3.5cm]{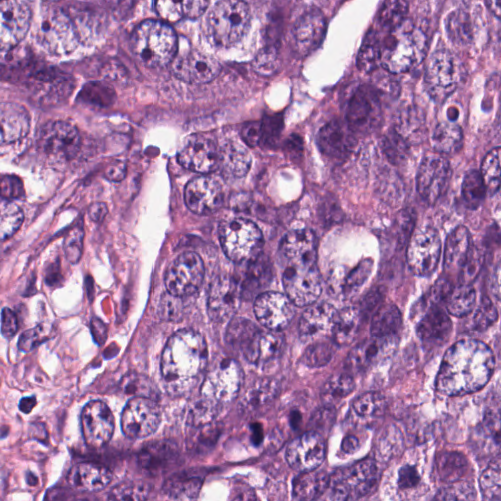}
      
      \caption{}
      \label{breast_cancer2}
    \end{subfigure}
  \end{center}
  \caption{(a)Colon cancer  (b)Prostate cancer  (c)Breast cancer (grade I) (d)Breast cancer(grade III)}
  \label{fig:image_sample}
\end{figure}

As an essential part of digital pathology, histopathology image analysis is playing increasingly important role in cancer diagnosis, which can provide direct and reliable evidence to diagnose the grade and type of cancer. This paper deals with nuclei segmentation, an important step in histopathological image analysis. The purpose of nuclei semgentation is not only counting the number of nuclei but also obtaining the detailed information of each nucleus. So unlike nuclei detection, here the outputs are the contour of each nucleus instead of only the position of their central points. Hence we can exactly extract each nucleus from the image and make it available for further analysis. For example, the features of the individual nucleus and the distribution of nuclei clusters can be used to grade and classify status of breast cancers \cite{nawaz2015computational,chen2015new}. Because of appearance variation such as color, shape, and texture, nuclei segmentation from histopathological images could be very challenging, as illustrated in Fig.\ref{fig:image_sample}, in which it is very challenging even for human to recognize and segment all nuclei within the images. Fig.\ref{colon} and Fig.\ref{prostate} illustrate two histopathological images from different organs.
Fig.\ref{breast_cancer1} and Fig.\ref{breast_cancer2} are two histopathological images from same organ but have different cancer grade.

   
The study of nuclei segmentation in histopathological images can be traced back to 10 years ago. A large number of methods have been proposed to pursue accurate segmentation on images of a variety of categories. The procedure of most traditional nuclei segmentation methods can be divided into two separate steps: first,detecting the nuclei and then obtaining each nucleus' contour. The detection step is expected to generate the area of nuclei or the seed of each nucleus. One popular and convenient method to detect the nucleus is intensity thresholding as used in Otsu's method\cite{otsu1975threshold}. However, it has an obvious limitation: it only works under the scenario that all the nuclei in the images have consistent intensity differences against the background. Another popular approach for nuclei detection is clustering including K-mean clusting\cite{filipczuk2011automatic}, Grab Cut\cite{rother2004grabcut} and etc. Furthermore, a few filtering based on methods have been proposed by utilizing the features of the nuclei\cite{al2010improved, veta2013automatic}. All of above methods have one common weakness: they are only effective for one or a few specific types of nuclei or images and are usually highly sensitive to manually set parameters. Since the appearances of nuclei are so diverse that we can hardly develop a single model or method suitable for all these different images. In recent years, supervised learning based approaches are becoming more and more attractive. They classify each pixel into one of two categories: nuclei or background\cite{mouelhi2013automatic,xu2016stacked,sirinukunwattana2016locality}. After the nuclei detection stage yields the nuclei area, the next step would be splitting the touching and overlapped nuclei areas. This could be achieved by methods such as bottleneck detection\cite{liao2016automatic} and ellipse fitting\cite{su2014automatic,kharma2007automatic}. If the seed of a nucleus is generated, its contour could be obtained by using marker controlled watershed\cite{veta2013automatic, qu2014two} or region growing\cite{xing2016automatic}.

Recently, deep learning based methods are becoming increasingly popular in image segmentation due to their dominating performance in many tasks of computer vision. They have significantly impacted all the research areas in computer vision such as object classification, object detection and segmentation. Since 2014, numerous convolutional neural network based image segmentation methods have been proposed. Long. et al firstly introduced
fully convolutional neural network (FCN)\cite{long2015fully} to semantic segmentation. Compared to prior models, it is demonstrated that the FCN algorithm is much more efficient and accurate. Converting fully connected layers into convolutional neural networks makes it possible to predict the heatmap of the objects in the image that needs to be segmented. U-net \cite{ronneberger2015u}, an FCN based network architecture won the Grand Challenge for Computer-Automated Detection of Caries in Bitewing Radiography at ISBI 2015. Later, a skip-architecture first introduced in residual networks \cite{he2016deep} is also applied to fuse different levels of semantic information. 


Inspired by the U-net algorithm\cite{ronneberger2015u}, we propose to apply the FCN network to the nuclei segmentation problem. Current deep learning methods for nuclei segmentation usually need complex post-processing procedure to obtain the final nuclei boundries. Naylor \cite{naylor2017nuclei} employs FCN to discriminate the nuclei from background and then applies the watershed method to split the nuclei. However the resulting boundaries are not accurate. Xing \cite{xing2016automatic} proposed a sophisticated shape deformation method to generate each nucleus's boundary.  Kumar\cite{kumar2017dataset} designed a CNN3 model to predict the nuclei and its boundary from the image. But a time consuming post-processing step is needed. Here we designed an end-to-end fully convolutional neural network architecture for nuclei segmentation. Unlike prior binary classifiers \cite{mouelhi2013automatic,xu2016stacked,sirinukunwattana2016locality}, which only discriminate nuclei against the background, our nuclei-boundary segmentation model predicts the nuclei and their contours at the same time. Due to the accurate prediction of nucleus and boundary in our approach, the final segmentation can be generated by a simple and fast post-processing procedure. To segment the whole-slide image, a pixel-wise segmentation strategy is necessary. However the border area of each patch cannot be predicted accurately because of lacking contextual information. A seamless patch extraction and assembling method is proposed to handle this problem.
The main contributions of this paper are as follows:

\begin{itemize}
  \item We propose a nuclei-boundary model to explicitly detect nuclei and their boundaries simultaneously from histopathology images. Detecting boundary is able to improve the accuracy of nuclei detection and help split the touched and overlapped nuclei. Given the raw segmentation results by our nuclei-boundary model, only a simple dilation operation and noise removing steps are needed to produce the final segmentation results. 
  \item We develop an effective approach to segment extra-large high-resolution images that U-net cannot handle due to limited GPU memory using a seamless patch-wise segmentation. A weighted loss map is utilized to train the model and a vote mechanism is used to assemble the patches.
  \item  Extensive studies on the effects of a variety of data augmentation methods for nuclei segmentation are provided.
  \item  We introduce four evaluation criteria for more accurate nuclei segmentation performance evaluation: missing detection rate, false detection rate, under-segmentation rate, and over-segmentation rate. They are designed to help the pathologist obtain more in-depth understanding of the performance of automatic segmentation methods and choose the right one for their specific application.
\end{itemize}

\section{Method}
\subsection{Overview}
\begin{figure}[!h]
	\begin{center}
		\includegraphics[width=0.9\linewidth]{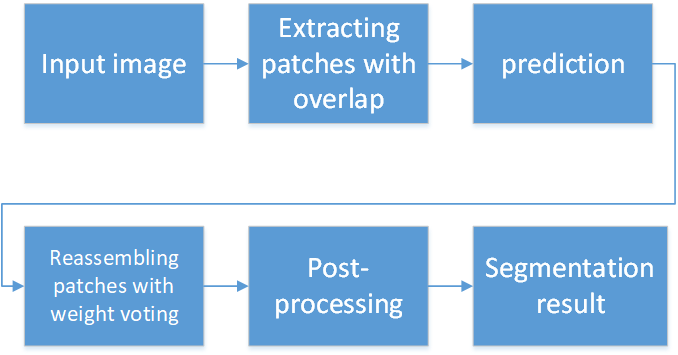}
	\end{center}
	\caption{The overview of segmenting nuclei on histopathological images.}
	\label{fig:overview}
\end{figure}

Our nuclei segmentation method adopts an end-to-end deep learning framework. The only preprocessing procedure is image color normalization. In the training phase, without extracting any features, even the H-channel, we directly apply the histopathology images in normalized RGB colors to the deep neural network to train the nucleus-boundary model. During the testing phase, the prediction result of raw normalized images yielded by the nucleus-boundary detector shows clear inside nuclei area and the boundaries. At last, we will obtain the  area of each nucleus via a simple, fast and parameterless post-processing procedure. Fig.\ref{fig:overview} shows the procedure to segment nuclei from color normalized images in our algorithm.

\subsection{Data Preprocessing}
H\&E stain is the most widely used stain protocol in medical diagnosis. Typically, the nuclei of cells are stained to blue by Haematoxylin while cytoplasm is colored to pink by Eosin. But in practice, the color of H\&E stained images could vary a lot due to variation in the H\&E reagents, staining process, scanner and the specialist who performs the staining, as shown in Fig.\ref{fig:image_sample}. A few H\&E stain normalization methods\cite{vahadane2015structure, khan2014nonlinear, macenko2009method} have been proposed to eliminate the negative interference caused by color variation. We tried two of them\cite{vahadane2015structure, macenko2009method} to normalize the raw H\&E stained images. For our segmentation algorithm, we did not find any considerable difference between these two normalization methods. Particularly, the result shown in experiment section \ref{Experiment} is generated based on the images normalized by the method in \cite{vahadane2015structure}. Given a target image, this method is able to convert one image's color into the target image's color space based on sparse non-negative matrix factorization(NMF). We choose one best stained H\&E image as the target and convert other images into its color space. According to the recommendation in \cite{vahadane2015structure}, the hyper-parameter $\lambda$ should be set between 0.01 and 0.1. In our experiment, $\lambda$ is set to 0.1.

Intuitively, the pure Haematoxylin-channel grayscale image would be much easier than RGB images to distinguish the foreground (nuclei) from the background (cytoplasm). A large number of nuclei segmentation methods\cite{cui2016self, qu2014two, wang2016automatic} employ some deconvolution algorithms to extract the H-channel from H\&E stained images. However, based on our experiments, we noticed that our deep fully convolutional neural network extracts the nuclei from raw RGB images better than from H-channel grayscale images. The reason would be that the H-channel might miss some information that might be helpful for distinguishing nuclei and the cytoplasm. Given well-labelled training images, the deep neural network can then learn the optimal way to extract the features that discriminate between each category of samples. So we skip extracting H-channel and directly apply the RGB color image as the input to  our deep neural network. 

\subsection{Nucleus-boundary model}
Traditional supervised nuclei segmentation methods usually apply a binary classifier to segment the nuclei area and the background area by classifying each pixel. These methods usually predict the category of the central pixel given a small patch. To segment the whole image, it needs to extract all the sliding windows(patches) with stride of 1 pixel and predict each of these patches. The most limitation of this strategy is high computational complexity. For example, if there is an image of size 1000X1000 pixel, this method needs to process one million sliding windows in order to segment this single image. Nevertheless, the typical whole slide of histopathology image may have billions of pixels, making it impossible to process it in an acceptable time using this strategy. Instead, our method is based on fully convolutional network (FCN) framework, which allows predicting the category of all the pixels of an image with only one pass. The input of the network is one image, the output is the estimated class map. 

\begin{figure*}[!ht]
  \begin{center}
    \includegraphics[width=.8\linewidth]{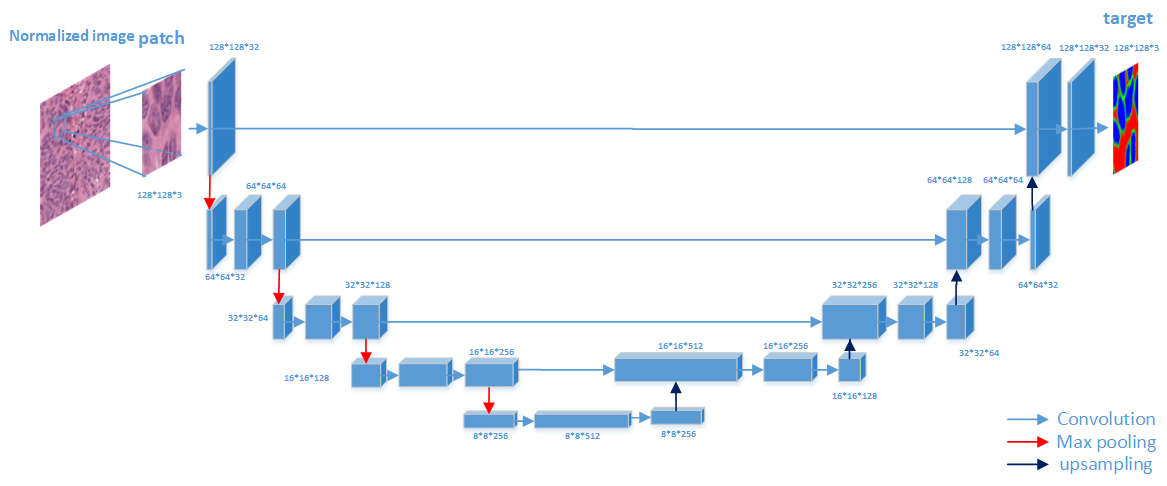}
  \end{center}
  \caption{The structure of our network. The size of each layer is shown in $height * width * channels$. The height and width of each layer are not fixed, which are determined by the size of input images. Here we assume the input image is of size $128 * 128$.}
  \label{fig:unet}
\end{figure*}

The task of nuclei segmentation can be roughly divided into two stages: the first stage is extracting the foreground(nuclei), the second stage is segmenting the connected foreground area into  separated nuclei and finding out the boundary of each nucleus. Our method intends to merge these two steps by extracting the nuclei and their edges at the same time. That is the reason why it is named "nuclei-boundary(NB) model". As shown in Fig.\ref{fig:unet}, the output of the NB model has three channels, each has the same height and width with the input image. Its values represent the probabilities of each pixel being $background$, $boundary$ or $inside$ class, respectively. The manual annotation for our segmentation problem is the boundary of each nucleus. A pixel belonging to the $boundary$ class means that it is on or inside an annotated boundary and within 2 pixel from the boundary. Pixels of the $inside$ class  are those that are inside annotated boundary but are not $boundary$ pixels. Correspondingly, the output can be regarded as an RGB image and the estimated maps of the $background$, $boundaries$ and $nuclei$ are represented by red, green and blue, respectively, as shown in Fig.\ref{fig:unet}. To generate the ternary mask for training, we apply a morphology operator to each nucleus to obtain the $inside$ pixels, and then subtract $inside$ pixels from the nucleus to get $boundary$ pixels.

\label{dense-unet}

\subsubsection{The architecture of our NB network}
Fig.\ref{fig:unet} shows the network architecture of our algorithm, which consists of a couple of encoding and decoding layers. The encoding layers are used to extract different levels of contextual feature maps. The decoding layers are designed to combine these feature maps produced by the encoding layers to generate the desired segmentation maps.
Due to the memory limitation of our GPU, the size of the input layer is set to 128X128 in our experiments. But we noticed that larger input layer may lead to better performance. 
The weight of each convolutional layer is initialized by glorot uniform\cite{glorot2010understanding} and bias is initialized to 0. The glorot uniform is defined as:
\begin{equation}	
W \sim  U\left[ \frac { -\sqrt { 6 }  }{ \sqrt { { n }_{ j } + { n }_{ j+1 } }  } ,\frac { \sqrt { 6 }  }{ \sqrt { { n }_{ j }+{ n }_{ j+1 } }  }  \right]
\label{eq:init}
\end{equation}
where $W$ means the initialized weight, $n_{j}$ means the size of the convolutional layer $j$. 

The scaled exponential linear units(SELUs) \cite{pmid29259059} activation function is used in all convolutional layers. SELUs is designed to make the forward neural network(FNN) to have self-normalizing capability\cite{klambauer2017self}. The FNN using SELUs are shown to be able to outputperform the ones using explicit normalization methods, such as batch normalization, layer normalization, and weight normalization. This is why our network does not have any normalization layers. The selu activation function is defined as:
\begin{equation*}
selu(x)=\lambda \begin{cases} x\quad \quad \quad \quad  \quad \quad if\quad x\quad <\quad 0 \\ \alpha { e }^{ x }\quad -\quad \alpha \quad  if\quad x\quad \le \quad 0\quad  \end{cases}
\label{eq:selu}
\end{equation*}
where 
$\lambda = 1.0507$ and $\alpha = 1.6733$. The padding property of each convolutional layer is the 'same' in order to ensure it keeps the same size with its previous layer. The size of all convolutional filters is $3 X 3$. Each convolutional layer is followed by a dropout layer with 0.2 drop rate. The network is trained by Adam optimizer\cite{kingma2014adam}. This stochastic optimization method is able to compute adaptive learning rate for each parameter. It automatically controls the learning rate along the training, so it is not necessary to manually set the momentum and decay. 


\subsubsection{Data Augmentation}
\label{data_augment}
Deep learning models often have millions of parameters so that it needs large-scale sample dataset to avoid the overfitting problem. In fact, the datasets of our nuclei segmentation task often contain only tens of images. Moreover, labeling an 1000*1000 image which contains hundreds of nuclei usually cost a specialist at least 5 hours. Hence it is impossible to manually label sufficient and  nuclei boundaries accurately for training deep learning models. Data augmentation is an essential approach to overcome the over-fitting problem caused by lacking samples. The training samples, \ie the patches, are randomly extracted from the H\&E stained images in the training datasets. Five augmentation techniques are used together in our experiments including random elastic transformation, rescale, affine transformation, shift, flip and rotate. Each training sample(one patch extracted from a whole image) as well as the corresponding target are processed by the data augmentation procedure.
Given a training sample, which is a RGB image $I$ with its corresponding ground truth $I_{gt}$, we transform $I$ to $I^{'}$ and $I_{gt}$ to $I_{gt}^{'}$. 
$I^{'}$ and $I_{gt}^{'}$ are the real input and target of the neural network. The rescaling factors are set as a random number between 0.5-1.5.
We employ Simard's method\cite{simard2003best} to do elastic transforming. Two hyper-parameters $\alpha$ and $\sigma$ need to be manually set to control how dramatic the original image is transformed. In our experiment, $\alpha$ is set to a random number between 100-200, $\sigma$ is set to 12. 

Besides transforming the input sample, it is necessary to do the same transformation on targets to maintain consistency. The one-hot encoding target consists of only binary values. However, the transformed target has some float-point numbers caused by bilinear interpolation we used for elastic transformation. They need to be binarized by the following rules:

Let the value of one pixel is $(t_{i}, t_{b}, t_{o})$, where $t_{i}$, $t_{b}$ and $t_{o}$ represents the label for $inside$, $boundary$ and $background$ respectively.

1. if $t_{b} > 0.5$, $t_{b} = 1$, else $t_{b} = 0$

2. if $t_{i} > 0 $and $t_{b} == 0$, $t_{i} = 1$, else $t_{i} = 0$

3. if $t_{i} == 1 $or $t_{b} == 1$, $t_{o} = 1$, else $t_{o} = 0$

\begin{figure}[!]
	\begin{center}
		\begin{subfigure}[t]{0.24\textwidth}
			\includegraphics[width=4.5cm]{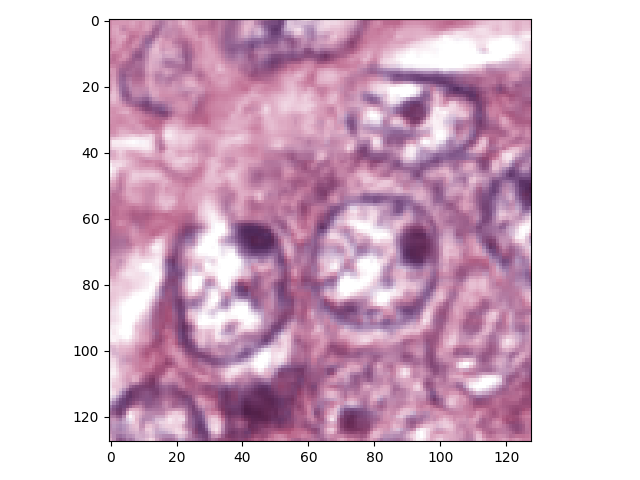}
			\caption{}
		\end{subfigure}
		\begin{subfigure}[t]{0.24\textwidth}
			\includegraphics[width=4.5cm]{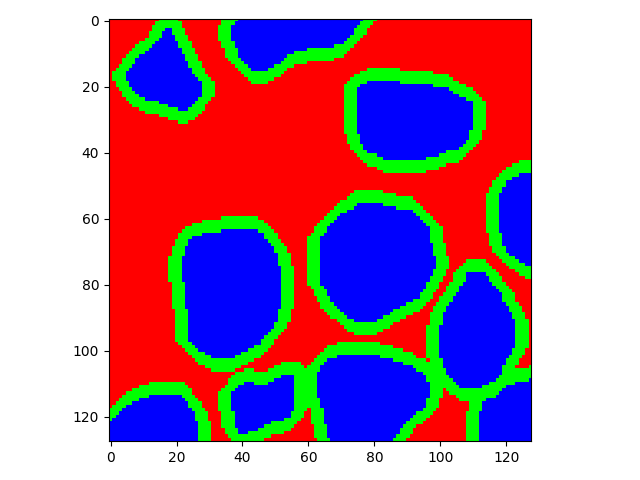}
			\caption{}
		\end{subfigure}
		
		\begin{subfigure}[t]{0.24\textwidth}
			\includegraphics[width=4.5cm]{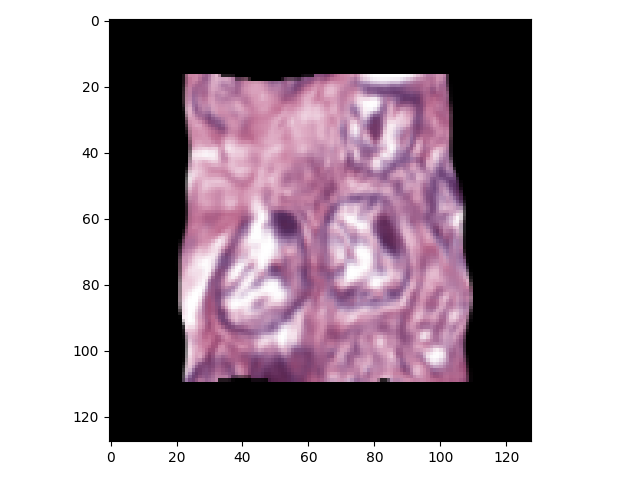}
			\caption{}
		\end{subfigure}
		\begin{subfigure}[t]{0.24\textwidth}
			\includegraphics[width=4.5cm]{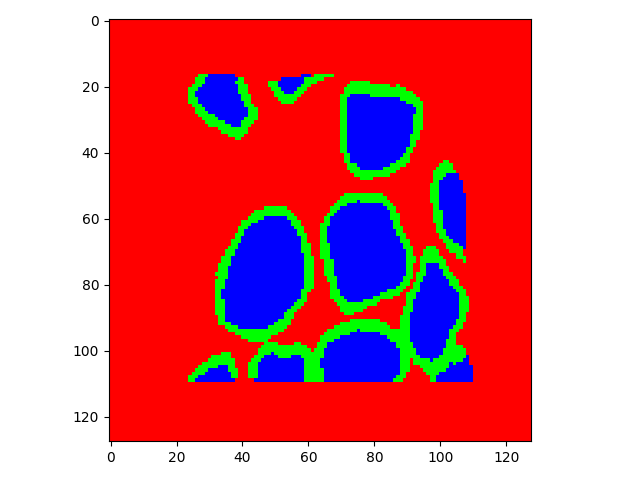}
			\caption{}
		\end{subfigure}
	\end{center}
	\caption{Example of data augmentation, (a) one patch extracted from a normalized image (b) corresponding ground truth of (a). (c) A training sample generated by data augmentation procedure based on patch(a). (d) the corresponding ground truth of (c).}
	\label{fig:augment}
\end{figure}

An example of data augmentation is illustrated in Fig.\ref{fig:augment}.

\subsubsection{Weighted loss}
 The U-net\cite{ronneberger2015u} model tends to predict the pixels with full context in the input image, which leads to generation of a smaller segmentation map than the input image. The border area of the input image is not predicted because of lacking enough context information. This strategy can solve the problem that the prediction of the border area is not accurate to some extent. One issue of this is that this U-net defines a fixed-size border area whose size is not changable without modifying the network structure.  However, in practice, the  border area size could vary in different histopathological images and it mainly depends on the size of the nuclei. Another limitation is that we have to do some cropping operation in neural network training to make the size of layers match each other, which might lose useful surrounding information.
 
 As a trade-off of these issues, we designed a weighted loss and a scheme for patch extraction and assembling to allow the neural network to predict an segmentation map of equal size without concerning the lack of context issue in the border area. 
 
 The model is trained by minimizing the categorical softmax cross-entropy loss between predictions and targets, which is described in eq.\ref{eq:loss}:
 
 \begin{equation}
L=\sum _{ i }^{  }{ \sum _{ j }^{  }{ { W }_{ i,j }log({ p }_{ t(i,j) }(i,j)) }  } 
\label{eq:loss}
 \end{equation}
 where 
 $t(i,j)$ denotes the true label of the pixel at (i,j) position, $p_{t(i,j)}(i,j)$ is the output of soft-max activation layer which indicates the probability of the pixel at (i,j) being $t(i,j)$. $W$ is the proposed weight map, which is defined as:

\begin{equation}
\begin{aligned}
&{ W }_{ i,j }={ \alpha \frac { { D }_{ i,j }^{ e } }{ { (D }_{ i,j }^{ c }+{ D }_{ i,j }^{ e }) }  }\\
&{ \alpha =\frac { h\cdot w }{ \sum _{ i=1 }^{ h }{ \sum _{ j=1 }^{ w }{ \frac { { D }_{ i,j }^{ e } }{ { D }_{ i,j }^{ c }+{ D }_{ i,j }^{ e } }  }  }  }  }  
\end{aligned}
\label{eq:loss_weight}
\end{equation}
 where $W_{i,j}$ is the weight of position i, j, 
, $D_{i,j}^{e}$ is the distance from border, $D_{i,j}^{c}$ denotes the distance from center.$h$ and $w$ are the height and width of the map, respectively.
\begin{figure}[hbp]
	\begin{center}
		\includegraphics[width=.4\linewidth]{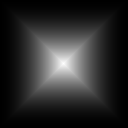}
	\end{center}
	\caption{The weighted loss map generated by Eq.\ref{eq:loss_weight}}
	\label{fig:weight}
\end{figure}

\subsubsection{Extra-large Image Segmentation Using Overlapped Patch Extraction and Assembling}
Current medical image segmentation algorithms based on U-net and its derivatives has an unsolved problem for segmenting extra-large high-resolution histopathological images: due to the limited memory of the GPU, it is possible to feed the whole slide image into the deep neural network. It has to be cut into patches and perform patch-wise training and prediction. However, there is no reported solution for deal with this issue. 

With close examination, we found the the main issue of U-net algorithm on patch-based segmentation is that the prediction at the border area is not accurate as demonstrated in \ref{fig:loss_weight_compare}. Here we propose an overlapped patch extraction and assembling method. The patches are extracted by sliding window with a stride. For assembling, a vote mechanism is applied to predict each pixel using
\begin{equation*}
P(i,j)\quad =\quad \frac { \sum _{ k }^{  }{ { W }_{ k(i,j) }p(k(i,j)) }  }{ \sum _{ k }^{  }{ { W }_{ k(i,j) } }  } 
\label{eq:assembling}
\end{equation*}
where $P(i,j)$ is the final prediction of the pixel at position (i,j) in an image. $k(i,j)$ means the position of 
it in the $kth$ patch. 

\subsubsection{Post-processing}

From Fig.\ref{fig:result_step}, we can see that the raw prediction results already show clear $inside$ nucleus areas and boundaries. Due to this reliable prediction results, we no longer need the complex region growing algorithms \cite{kumar2017dataset, xing2016automatic} and splitting algorithms \cite{wang2016automatic} to extract the final segmented areas. These methods usually strongly rely on manual parameter tuning to get good performance and is computationally demanding. Instead, we use a parameter-free postprocessing procedure that runs in a negligibly short time. Since our NB model detects both $inside$ and $boundary$ classes, all we need is the $inside$ class map. Then the $inside$ class map is transformed to a binary map using a constant threshold $0.5$. 
In this way, each connected component in the binary image indicates the $inside$ area of one nucleus. At the end, in order to recover the shape, we can simply apply the dilation operation to each connected component.

\section{Experiment}
\label{Experiment}
\subsection{Evaluation criteria}
Two level of criteria are usually used to measure the performance of nuclei segmentation methods: one is object-level criteria, another is pixel-level criteria.
The most common object-level criteria for object detection tasks include $precision$, $recall$, $F1 score$. $precision$ is defined as:
\begin{equation*}
precision\quad =\quad \frac { TP }{ TP\quad +\quad FP } 
\end{equation*}
$recall$ is defined as:
\begin{equation*}
recall\quad =\quad \frac { TP }{ FN\quad +\quad TP } 
\end{equation*}
$F1 score$ considers both of the $precision$ and $recall$, as shown in following equation. 
\begin{equation*}
F1=2\cdot \frac { precision\cdot recall }{ precision+recall } 
\end{equation*}
where the $TP$ is true positives, $FP$ means false positives and $FN$ means false negatives. Given a manually labelled ground truth nucleus $T_{i}$, if there is one nucleus $P_{j}$ in automatic segmentation result that matches $T_{i}$, $P_{j}$ can be counted as one $TP$. 

$F1 \quad  score$ is the harmonic average of $precision$ and $recall$ and its value is in the range of [0,1]. 

We noticed that $FN$ can be caused by two different types of errors: one is miss-detection(nuclei is predicted as cytoplasm), another is under-segmentation(Multiple ground truth nuclei are detected as one nucleus, hence only one of these nuclei ground truth nucleus has corresponding detected nucleus.).Similarly, $FP$ consists two types of errors: one is false detection (Cytoplasm is detected as nuclei), another is over-segmentation(One ground-truth nucleus is segmented into several nuclei. Each of them is a part of the ground truth nucleus and at most only one among them can be considered as the corresponding detected nucleus). 
Let us think about this situation: one segmentation method is weak on discriminating the nuclei and cytoplasm while another one is weak on splitting the nuclei area. 
But they may have similar $precision$ and $recall$, even $F1 score$. Apparently, $precision$, $recall$, $F1 score$ and their combination fail to differentiate the performance of these two segmentation methods. To handle this issue, we introduce four new criteria to evaluate automatic nuclei segmentation methods: missing detection rate(MDR), false detection rate(FDR), under-segmentation rate(USR), over-segmentation rate(OSR), as shown in Eq. \ref{eq:1}.
\begin{align}
&MDR=\frac { MD }{ FN\quad +\quad TP } \nonumber\\
&FDR=\frac { FD }{ TP\quad + \quad FP } \nonumber\\
&USR=\frac { US }{ P }\nonumber\\
&OSR=\frac { OS }{ S } \label{eq:1}
\end{align}
where $MD$ is the number of missing detections, $FD$ indicates the number of false detections, $US$ means the number of nuclei which are not detected caused by undersegmentation. $P$ is the number of ground truth nuclei in the region of $TP$, which can be defined as $Fn + TP - MD$. $OS$ means the number of false positives caused by oversegmentation 
and $S$ means the number of segmented nuclei in the region of $TP$'s corresponding ground truth nuclei, which can be defined as $FP + TP + FD$.
The combination of $MDR$ and $FDR$ measures the capacity of discriminating the nuclei and cytoplasm while the combination of $USR$ and $OSR$ measures the performance of handling overlapped nuclei area. On the other hand, $recall$ value is negatively correlated with $MD$ and $USR$ while $precision$ is negatively correlated with $FDR$ and $OSR$. These four criteria are able to help pathologists to select proper automatic segmentation methods for specific tasks.

The pixel-level criteria are used to measure the accuracy of a segmentation algorithms in predicting the shape and size of the detected nuclei. The most essential one is Dice's coefficient, which is defined as:
\begin{equation}
D(X,Y)=2\frac { \left| X\cap Y \right|  }{ \left| X \right| +\left| Y \right|  } 
\end{equation}
where $X$ indicates a manual segmentation and $Y$ means its corresponding automatic segmentation. That is, a manual segmentation is considered as a FP if there is no corresponding automatic segmentation with a Dice coefficient of at least 0.2.  

\subsubsection{Datasets}
We evaluate the performance our method on three public available nuclei segmentation datasets. One is a multiple organ H\&E stained image dataset\cite{kumar2017dataset}(MOD). It consists of 30 images which were captured from 7 organs: breast, liver, kidney, prostate, bladder, colon and stomach. The resolution of each image is 1000X1000. Totally, about 21,000 nuclear boundaries are manually annotated. These 30 images are split into two subsets: the training set with 16 images composed of 4 from breast, 4 from liver, 4 from kidney and 4 from prostate and the test set with 14 images composed of 2 images from each organ.

The second dataset is the breast cancer histopathology image dataset(BCD). It contains two subsets: subsetA and subsetB. SubsetA includes 21 images and subsetB has 18 images. In \cite{veta2013automatic}, SubsetA is used to tune the parameters. In a similar way, we utilize subsetA as the training set and subsetB as the test set. Since one image may contains thousands of nuclei, it is impractical to manually label all the training images. We randomly select five images from subsetA and crop a 1000*1000 subimage from each of them to build the training set. It is manually annotated under the supervision of a specialist.

The third one is also a breast cancer image dataset(BNS)\cite{naylor2017nuclei}. It is composed of 33 H\&E stained images of size 512X512 from 7 triple negative breast cancer patients. There are totally 2754 manually annotated nuclei.
\subsection{Experiment result}
\begin{figure*}[htbp]
	\begin{center}
		\begin{subfigure}[t]{0.24\textwidth}
			\includegraphics[width=4cm]{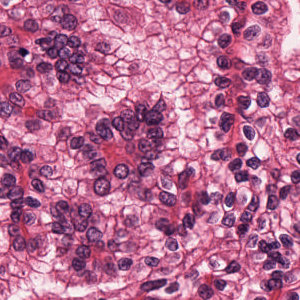}
		\end{subfigure}
		\begin{subfigure}[t]{0.24\textwidth}
			\includegraphics[width=4cm]{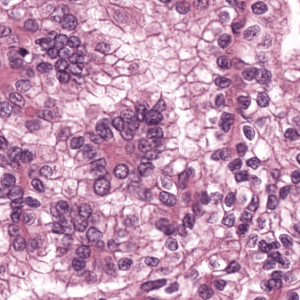}
		\end{subfigure}
		\begin{subfigure}[t]{0.24\textwidth}
			\includegraphics[width=4cm]{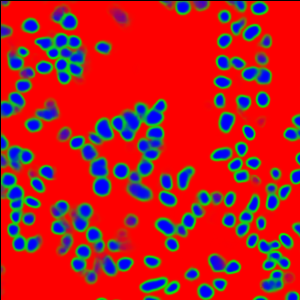}
		\end{subfigure}
		\begin{subfigure}[t]{0.24\textwidth}
			\includegraphics[width=4cm]{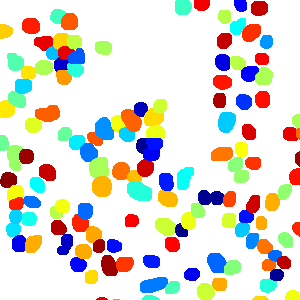}
		\end{subfigure}

     \hspace{3mm}

		~
		\begin{subfigure}[t]{0.24\textwidth}
			\includegraphics[width=4cm]{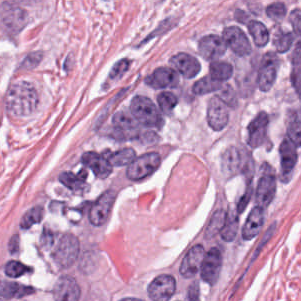}
		\end{subfigure}
		\begin{subfigure}[t]{0.24\textwidth}
			\includegraphics[width=4cm]{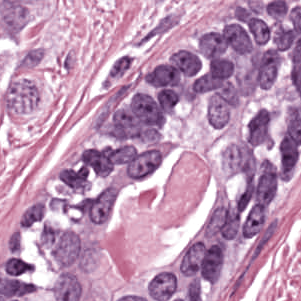}
		\end{subfigure}
		\begin{subfigure}[t]{0.24\textwidth}
			\includegraphics[width=4cm]{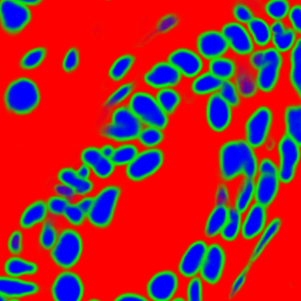}
		\end{subfigure}
		\begin{subfigure}[t]{0.24\textwidth}
			\includegraphics[width=4cm]{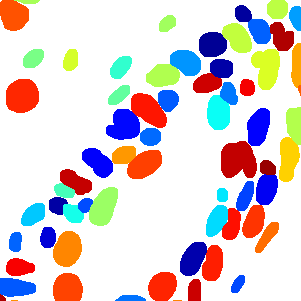}
		\end{subfigure}

     \hspace{3mm}

		~
		\begin{subfigure}[t]{0.24\textwidth}
			\includegraphics[width=4cm]{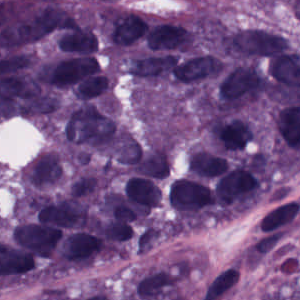}
			\caption{}
		\end{subfigure}
		\begin{subfigure}[t]{0.24\textwidth}
			\includegraphics[width=4cm]{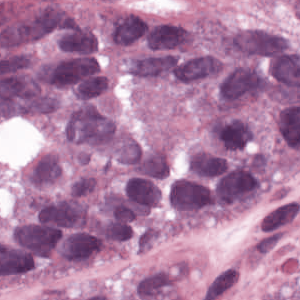}
			\caption{}
		\end{subfigure}
		\begin{subfigure}[t]{0.24\textwidth}
			\includegraphics[width=4cm]{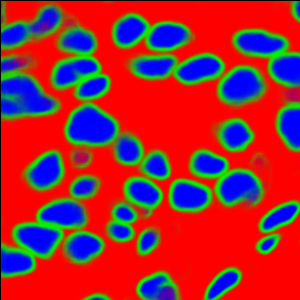}
			\caption{}
		\end{subfigure}
		\begin{subfigure}[t]{0.24\textwidth}
			\includegraphics[width=4cm]{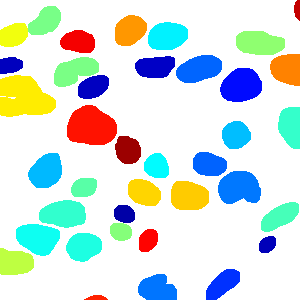}
			\caption{}
		\end{subfigure}
		\caption{(a) examples of original histopathology images; (b) corresponding images after color normalization. (c) raw segmentation results by our algorithm. (d) final segmentation result.}
		\label{fig:result_step}
	\end{center}
	
\end{figure*}

Figure \ref{fig:result_step} shows how our method segments the nuclei step by step. The color variety is well controlled by the color normalization procedure. The prediction result shows clear nuclear areas and nucleus boundaries. In the final segmentation result and ground truth image, each nucleus is represented by a different color. 

\begin{figure*}[htbp]
	\begin{center}
		\begin{subfigure}[t]{0.24\textwidth}
			\includegraphics[width=4cm]{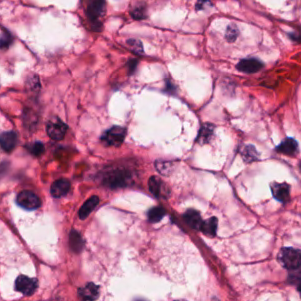}
			
		\end{subfigure}
		\begin{subfigure}[t]{0.24\textwidth}
			\includegraphics[width=4cm]{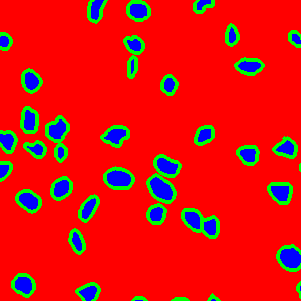}
			
		\end{subfigure}
		\begin{subfigure}[t]{0.24\textwidth}
			\includegraphics[width=4cm]{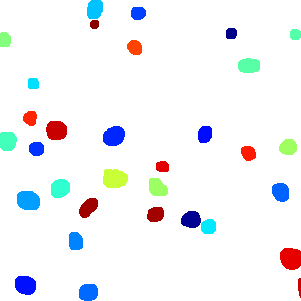}
			
		\end{subfigure}
		\begin{subfigure}[t]{0.24\textwidth}
			\includegraphics[width=4cm]{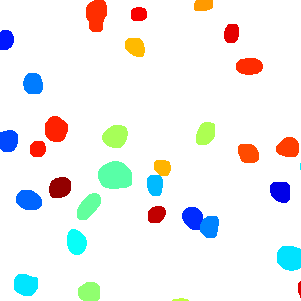}
			
		\end{subfigure}

    \hspace{3mm}

		~
		\begin{subfigure}[t]{0.24\textwidth}
			\includegraphics[width=4cm]{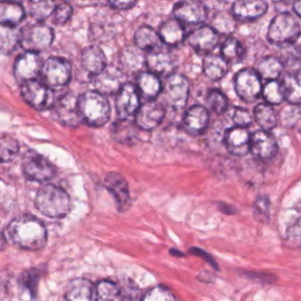}
			
		\end{subfigure}
		\begin{subfigure}[t]{0.24\textwidth}
			\includegraphics[width=4cm]{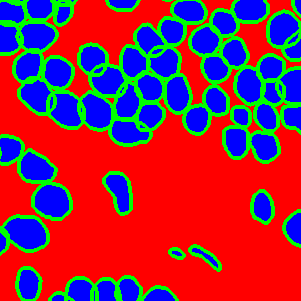}
			
		\end{subfigure}
		\begin{subfigure}[t]{0.24\textwidth}
			\includegraphics[width=4cm]{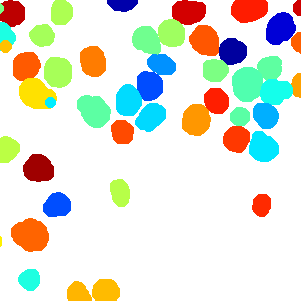}
			
		\end{subfigure}
		\begin{subfigure}[t]{0.24\textwidth}
			\includegraphics[width=4cm]{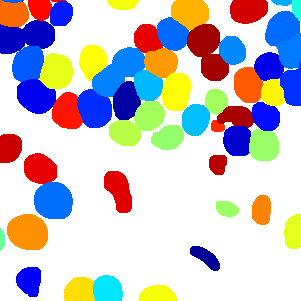}
			
		\end{subfigure}

    \hspace{3mm}

		~
		\begin{subfigure}[t]{0.24\textwidth}
			\includegraphics[width=4cm]{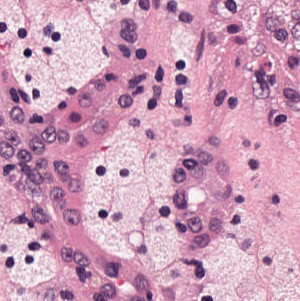}
			\caption{}
		\end{subfigure}
		\begin{subfigure}[t]{0.24\textwidth}
			\includegraphics[width=4cm]{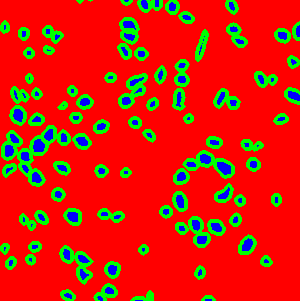}
			\caption{}
		\end{subfigure}
		\begin{subfigure}[t]{0.24\textwidth}
			\includegraphics[width=4cm]{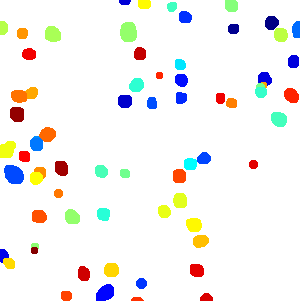}
			\caption{}
		\end{subfigure}
		\begin{subfigure}[t]{0.24\textwidth}
			\includegraphics[width=4cm]{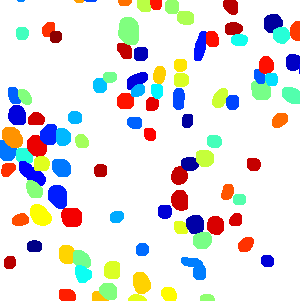}
			\caption{}
		\end{subfigure}
		\caption{The comparison between our method and CNN3\cite{kumar2017dataset}. (a): raw images; (b):ground truth; (c): CNN3 results; (d): our results }
		\label{fig:compare_TMI}
	\end{center}
\end{figure*}

First, We test our method on the MOD dataset. Unfortunately, the dataset publicly provided online doesn't explicitly divide the whole dataset into the training set and test set. We do not know which image belongs to the training set exactly as introduced in their paper \cite{kumar2017dataset}. To make a fair comparison, we randomly select 16 images from breast, liver, kidney and prostate. Then we combine the remaining 8 images of these four types and the 6 images from bladder, colon and stomach to build the test images. 12000 patches are randomly extracted from 12 training images to train our model. To eliminate the bias caused by random selection, 5 different training sets and the corresponding test sets are randomly generated. Then the model is trained and tested on the 5 pairs of training set and test set separately. All of the models are trained for 300 epoch in 7.5 hours. For testing, the stride of overlapped patch extraction is set to 64. The quantitative comparison is listed in Table  \ref{table:1}, which demonstrates that our method outperforms the state-of-the-art method CNN3 as reported in \cite{kumar2017dataset} in terms of both F1 score and Dice's Coefficient. Moreover, it shows that the under-segmentation error is much more significant than over-segmentation error and it achieves a balance between the false detection error and missing detection error.
Figure \ref{fig:compare_TMI} shows a visual comparison between our method and \cite{kumar2017dataset}. As shown in the sample images, our segmentation result has fewer false negatives and higher accuracy in terms of nuclei boundaries than \cite{kumar2017dataset}. Our method is not only more accurate but also much faster. It takes about 5 seconds to predict a 1000 * 1000 image by one Nvidia Titan X GPU and the time used for post-processing is less than 0.1 seconds. Given the same hardware environment and test images, \cite{kumar2017dataset} takes about 4 minutes to predict one image and 80 seconds to do the post-processing.
Additionally, a 10-folder cross-validation is performed to validate our method. The result is shown in Table \ref{table:1} NB model *.
 
\begin{table*}[t]
	\begin{center}
		\begin{tabular}{|c|c|c|c|c|c|c|c|c|}
			\hline
			methods &  precision & recall & F1 & Dice's Coefficient & 
			MDR & FDR & USR & OSR \\
			\hline
			\hline
			CNN3 \cite{kumar2017dataset} & - & - & 0.827 & 0.762 & - & - & - & - \\
			\hline
			NB model 1(our method) &  0.813 & 0.914  & 0.854  &  0.812 & 0.09 & 0.09 & 0.09 & 0.01\\
			\hline
			NB model 2  & 0.861 & 0.856 & 0.846 & 0.808 & 0.05 & 0.13 & 0.08 & 0.03\\
			\hline
			NB model 3 & 0.880 & 0.864 & 0.854 & 0.818 & 0.07 & 0.11 & 0.05 & 0.03\\
			\hline
			NB model 4 & 0.812 & 0.925 & 0.861 & 0.805 & 0.09 & 0.07 & 0.09 & 0.01\\
			\hline
			NB model 5 & 0.814 & 0.910 & 0.846 & 0.803 & 0.10 & 0.08 & 0.10 & 0.01\\
			\hline
			NB model * & 0.845 & 0.892 & 0.850 & 0.81 & 0.06 & 0.11 & 0.02 & 0.08\\
			\hline
		\end{tabular}
	\end{center}
	\caption{Quantitative comparison results of segmentation performance on MOD dataset. }
	\label{table:1}
\end{table*}

\begin{figure}[htbp]
	\begin{center}
		\begin{subfigure}[t]{0.15\textwidth}
			\includegraphics[width=2.5cm]{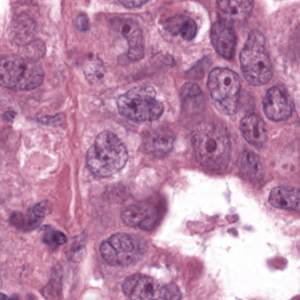}
		\end{subfigure}
		\begin{subfigure}[t]{0.15\textwidth}
			\includegraphics[width=2.5cm]{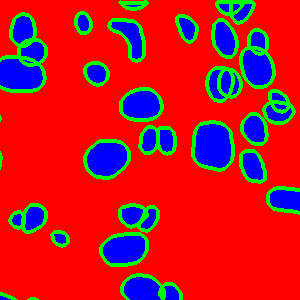}
			\caption{}
		\end{subfigure}
		\begin{subfigure}[t]{0.15\textwidth}
			\includegraphics[width=2.5cm]{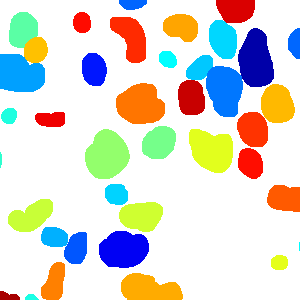}
		\end{subfigure}
		~
		\begin{subfigure}[t]{0.15\textwidth}
			\includegraphics[width=2.5cm]{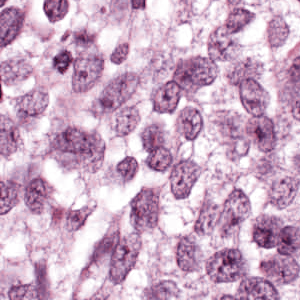}
		\end{subfigure}
		\begin{subfigure}[t]{0.15\textwidth}
			\includegraphics[width=2.5cm]{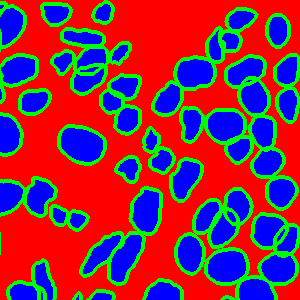}
			\caption{}
		\end{subfigure}
		\begin{subfigure}[t]{0.15\textwidth}
			\includegraphics[width=2.5cm]{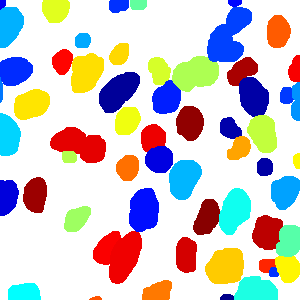}
		\end{subfigure}
		\caption{Cropped portions of two images. (a) precision = 0.76 recall = 0.83, OSR = 0.05 USR = 0.15 MDR = 0.02 FDR = 0.20 (b) precision = 0.78 recall = 0.83 OSR = 0.13 USR = 0.05 MDR = 0.12 FDR = 0.10}
		\label{fig:OSR}
	\end{center}
\end{figure}

 To show the benefit of our proposed evaluation metrics for nuclei segmentation, we compared the performance of our algorithm and the baseline CNN3 over two images with similar precision and recall, but different segmentation quality. As shown in Fig. \ref{fig:OSR}, the CNN3 algorithm got similar precision and recall scores on these two images. 
From our proposed criteria, we can find that the segmentation error on the first image is mainly caused by under-segmentation and false detections while that it is mainly caused by oversegmentation, missing detection and false detection in the second image. This observation can be verified by the sample segmentation result. 

\begin{figure*}[htbp]
	\begin{center}
		\begin{subfigure}[t]{0.32\textwidth}
			\includegraphics[width=5.5cm]{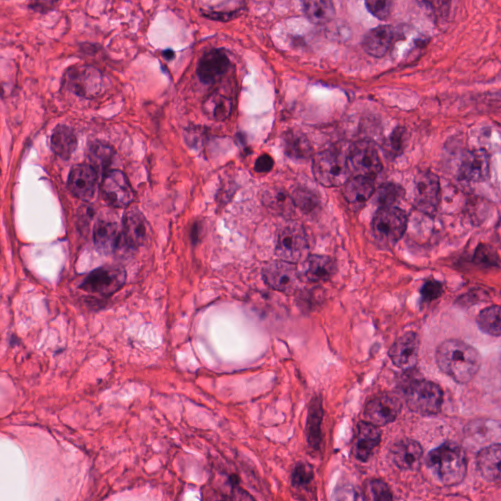}
		\end{subfigure}
		\begin{subfigure}[t]{0.32\textwidth}
			\includegraphics[width=5.5cm]{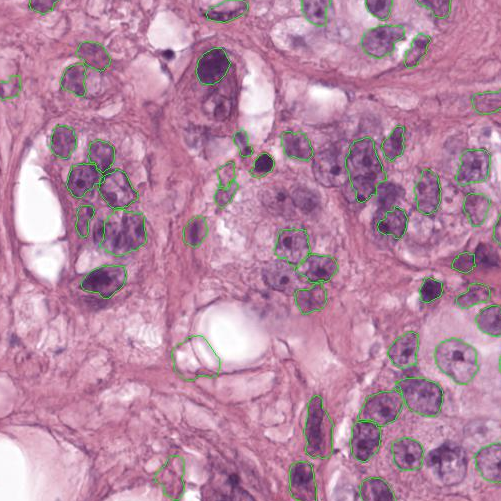}
		\end{subfigure}
		\begin{subfigure}[t]{0.32\textwidth}
			\includegraphics[width=5.5cm]{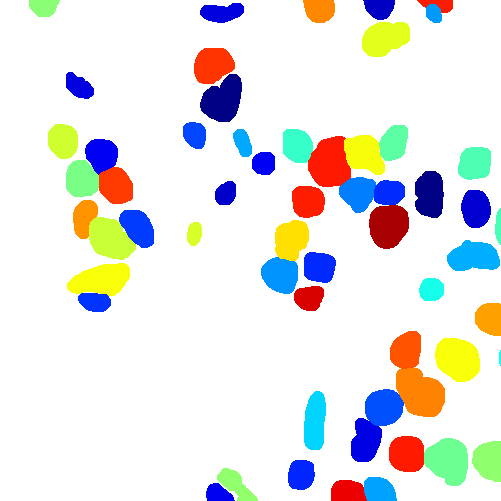}
		\end{subfigure}

     \hspace{3mm}

		~
		\begin{subfigure}[t]{0.32\textwidth}
			\includegraphics[width=5.5cm]{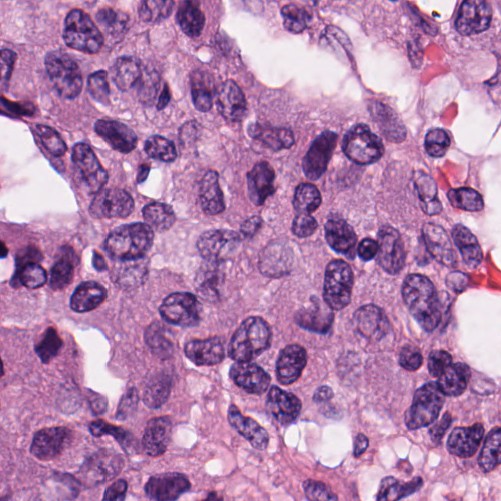}
		\end{subfigure}
		\begin{subfigure}[t]{0.32\textwidth}
			\includegraphics[width=5.5cm]{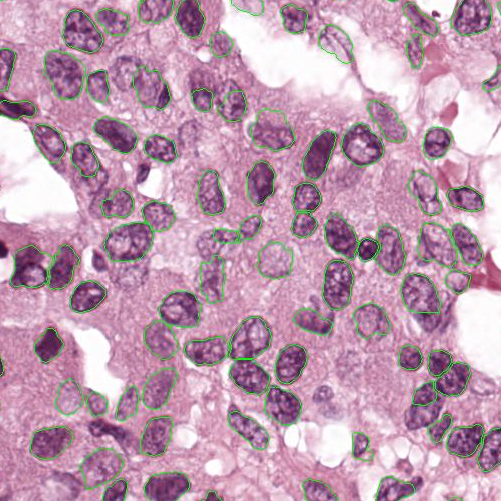}
		\end{subfigure}
		\begin{subfigure}[t]{0.32\textwidth}
			\includegraphics[width=5.5cm]{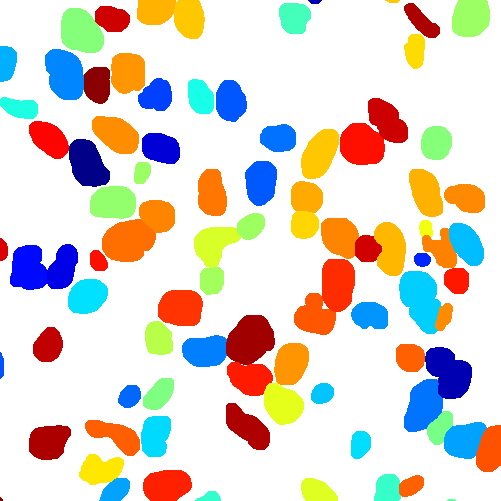}
		\end{subfigure}

     \hspace{3mm}

		~
		\begin{subfigure}[t]{0.32\textwidth}
			\includegraphics[width=5.5cm]{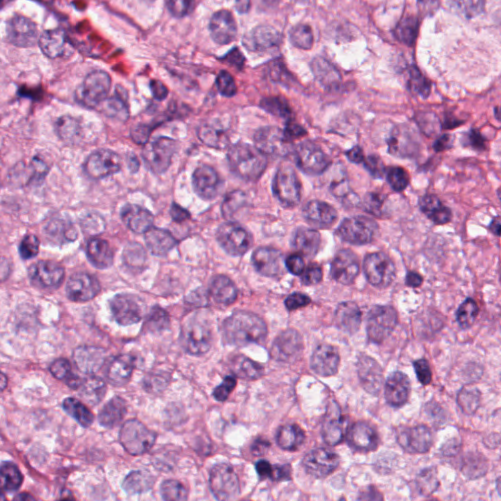}
		\end{subfigure}
		\begin{subfigure}[t]{0.32\textwidth}
			\includegraphics[width=5.5cm]{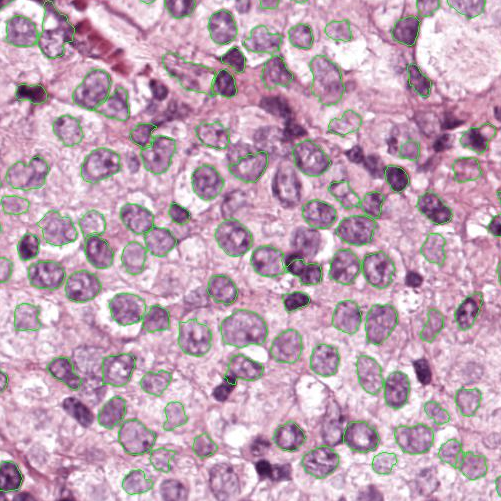}
		\end{subfigure}
		\begin{subfigure}[t]{0.32\textwidth}
			\includegraphics[width=5.5cm]{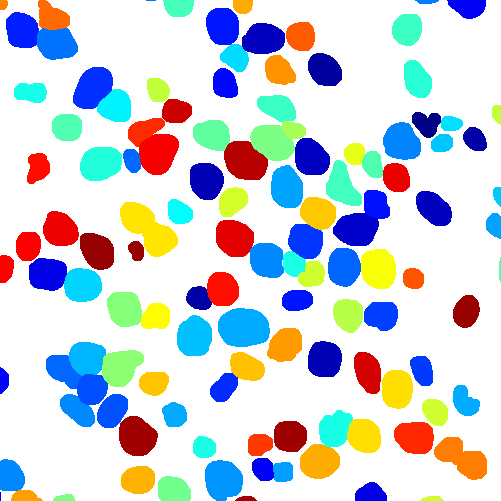}
		\end{subfigure}

     \hspace{3mm}

		~
		\begin{subfigure}[t]{0.32\textwidth}
			\includegraphics[width=5.5cm]{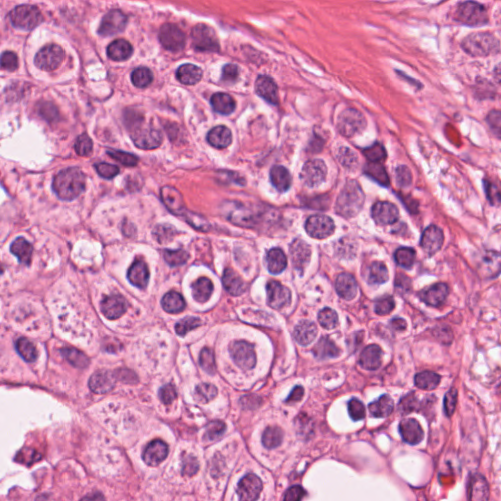}
			\caption{}
		\end{subfigure}
		\begin{subfigure}[t]{0.32\textwidth}
			\includegraphics[width=5.5cm]{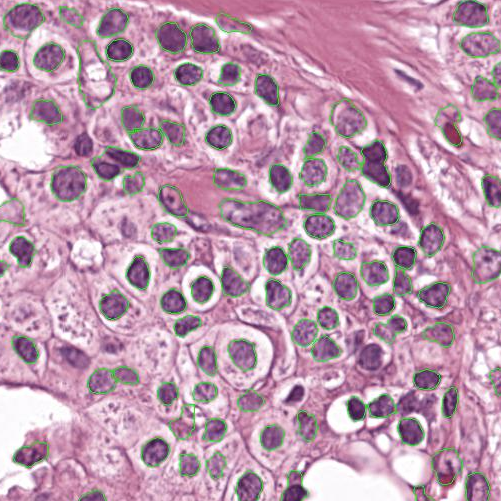}
			\caption{}
		\end{subfigure}
		\begin{subfigure}[t]{0.32\textwidth}
			\includegraphics[width=5.5cm]{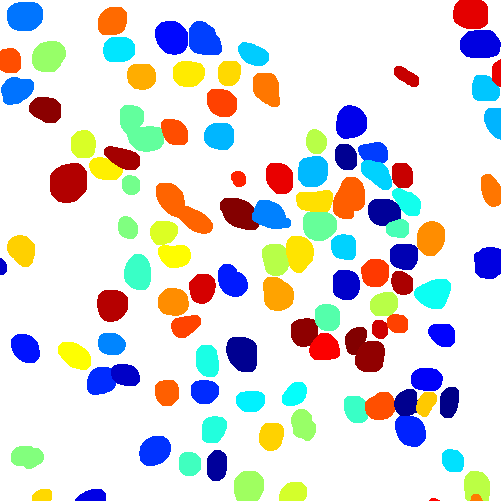}
			\caption{}
		\end{subfigure}
		\caption{Nuclei segmentation result over the BCD dataset. (a) two breast cancer image samples. (b) automatic segmentation result of \cite{veta2013automatic}. (c) result of our method.}
		\label{fig:result1}
	\end{center}
\end{figure*}

Second, we test our method on the BCD dataset.The manually labeled training set consists of five 1000*1000 images. Instead of training the models from random initialization, we use the training data to fine-tune the network model trained on the MOD dataset. Thus the model would adjust to a new dataset with much shorter time by training on a limited training set for a small number of epochs. In this experiment, only 2000 patches are extracted to fine-tune the pre-trained model. It takes about 10 seconds to train one epoch and the training is terminated after 70 epochs.
Figure \ref{fig:result1} shows the visual comparison between our algorithm and algorithm in \cite{veta2013automatic} in terms of segmentation results. 
At last, we follow the same strategy in \cite{naylor2017nuclei} to validate our method. The strategy is called leave-one-patient-out cross-validation. That is every time we train the model on 6 patient and use the rest one for validation.  
Table \ref{table:2} shows that our method outperforms the state-of-the-art  breast cancer nuclei segmentation method by a large margin in terms of precision, recall and F1 score.
\begin{table}[t]
	\begin{center}
		\begin{tabular}{|c|c|c|c|c|c|}
			\hline
			dataset & method & precision & recall & F1 & DC\\
			\hline   
			\hline
			BCD & Veta's method\cite{veta2013automatic} & 0.863 & 0.886 & 0.874 & ~0.88  \\
			\hline
			& TV-MRF-BP \cite{paramanandam2016automated} & 0.801 & 0.823  & 0.811 & 0.84  \\
			\hline
			& NB model  & 0.942 & 0.915 & 0.923 & 0.862 \\
			\hline
			BNC & FCN\cite{long2015fully} & 0.823 & 0.752 & 0.763 & - \\
			\hline
			& DeconvNet\cite{noh2015learning} & 0.864 & 0.773 & 0.805 &-\\
			\hline
			& Ensemble\cite{naylor2017nuclei} & 0.741 & 0.9 & 0.802& -\\
			\hline
			& NB model & 0.920 & 0.7835 & 0.84 & 0.83\\
			\hline
		\end{tabular}
	\end{center}
	\caption{Quantitative comparison of segmentation performance on the BCD dataset}
	\label{table:2}
\end{table}

\subsection{Discussion}
\subsubsection{Data augmentation for fully convolutional networks }
Data augmentation is a widely used technique to handle the overfitting issue caused by limited training samples. In image segmentation tasks, one can generate more images from one image using image transformation methods. The most common methods include rotation, flipping, shifting and rescaling. Elastic deformation transform, a higher level transformation method, is also employed in some image segmentation works. Ronneberger \textit{et al.} \cite{ronneberger2015u} claim that elastic deformation is the key method to do data augmentation for a segmentation network with very limited annotated images. 

\begin{figure}[!ht]
	\begin{center}
			\begin{subfigure}[t]{0.40\textwidth}
				\includegraphics[width=6cm]{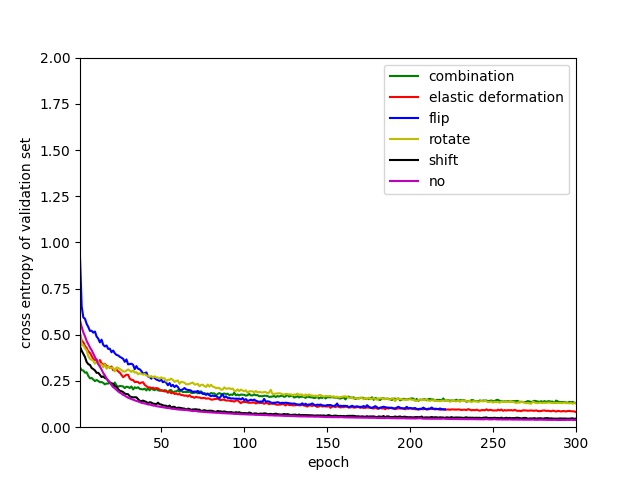}
				\caption{}
			\end{subfigure}
			\begin{subfigure}[t]{0.40\textwidth}
				\includegraphics[width=6cm]{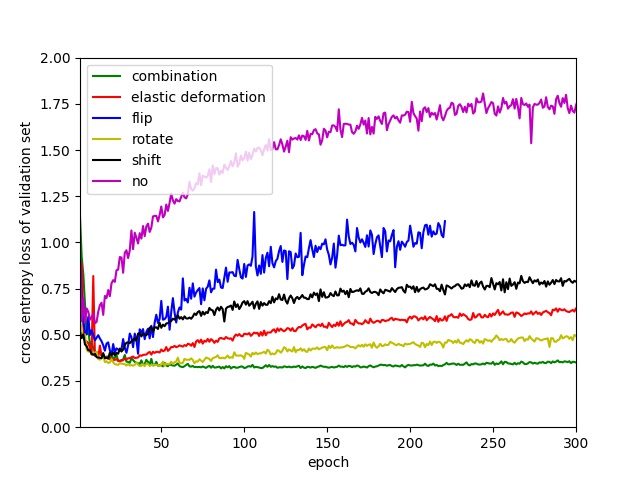}
				\caption{}
			\end{subfigure}
	\end{center}
	\caption{(a) shows how the training loss changes during training. (b) indicates the validation loss.}.
	\label{fig:overfitting}
\end{figure}

However, to the best of our knowledge, there is no systematic study of the effectiveness of these image transformation methods for nuclei segmentation using a fully convolutional network. We compare different training processes using rotation, flipping, shifting, rescaling and elastic deformation transform to augment the training data. To make fair comparisons, we let the training set and validation set have similar appearances by splitting each whole image into two sub-images and placing one in the training set and another one in the validation set. We randomly extract 6000 patches from the training set to train our neural networks and 6000 patches from the validation set for validation. The setting of these transformation methods is same with those reported in section \ref{data_augment}. The comparison is shown in Fig.\ref{fig:overfitting}. 'no' means don't apply data augmentation. 'combination' means data augmentation is performed by combining elastic deformation, flip, rotate, shift and rescale. It is very clear that without data augmentation, the network has severe overfitting issue, validation loss starts to increase rapidly from epoch 5. Unexpectedly, rotating rather than elastic deformation has achieved the best performance in performance improvement. But only rotating operation still cannot prevent the overfitting. One has to combine all of these transform methods together to do data augumentation to get good performance as done in this paper.

\subsubsection{Nuclei Segmentation on Extra-large Images}
\begin{figure}[!h]
	\begin{subfigure}[t]{0.24\textwidth}
		\includegraphics[width=4cm]{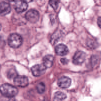}
		\caption{}
	\end{subfigure}
	\begin{subfigure}[t]{0.24\textwidth}
		\includegraphics[width=4cm]{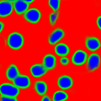}
		\caption{}
		\label{fig:pred}
	\end{subfigure}
	~
	\begin{subfigure}[t]{0.24\textwidth}
		\includegraphics[width=4cm]{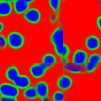}
		\caption{}
	\end{subfigure}
		\begin{subfigure}[t]{0.24\textwidth}
			\includegraphics[width=4cm]{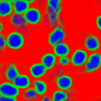}
			\caption{}
			\label{fig:loss}
		\end{subfigure}
	\caption{(a) shows an H\&E stained image. (b) shows the raw segmentation results of our method. (c) shows the prediction result without applying weight map and overlapped patch extraction and assembling. (d) shows the prediction results using overlapped patch extraction and assembling but without weight map.}  
	\label{fig:loss_weight_compare}
\end{figure}

To evaluate the effectiveness of the proposed weight map and overlapped patch extraction and assembling method for extra-large image segmentation, we compared the segmentation results with and without the proposed method in Fig. \ref{fig:loss_weight_compare}.  We can see that the raw segmentation results without using those two techniques contain obvious seams between the patches. It also demonstrates that the predictions in the border area is not accurate. As shown in Fig. \ref{fig:loss}, if we employ the overlapped patch extraction and assembling but without the weight map (which means all the pixels in a patch have the same weight) the segmentation result still shows noticeable seams. Fig.\ref{fig:pred} and Fig. \ref{fig:loss} has the same stride, which is 64.

\subsubsection{NB model versus the mixed nucleus model + boundary model}
\label{double}

\begin{figure}[!ht]
  \begin{center}
    \begin{subfigure}[t]{0.15\textwidth}
      \includegraphics[width=2.5cm]{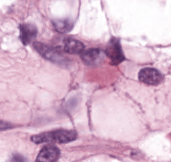}
    \end{subfigure}
    \begin{subfigure}[t]{0.15\textwidth}
      \includegraphics[width=2.5cm]{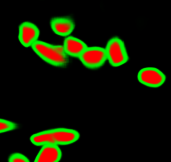}
    \end{subfigure}
    \begin{subfigure}[t]{0.15\textwidth}
      \includegraphics[width=2.5cm]{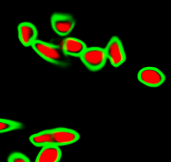}
    \end{subfigure}
    
    \hspace{3mm}
    
    \begin{subfigure}[t]{0.15\textwidth}
      \includegraphics[width=2.5cm]{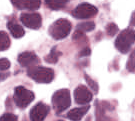}
    \end{subfigure}
    \begin{subfigure}[t]{0.15\textwidth}
      \includegraphics[width=2.5cm]{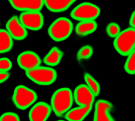}
    \end{subfigure}
    \begin{subfigure}[t]{0.15\textwidth}
      \includegraphics[width=2.5cm]{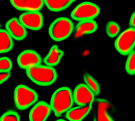}
    \end{subfigure}
    \caption{The comparison of NB model against the mixed nucleus model and boundary model. First column shows the histophysiological images. Second column shows estimated nuclei and boundaries using our NB model. Third column shows the estimated result generated by the mixed nucleus model and boundary model. }
    \label{fig:three_two}
  \end{center}
\end{figure}

An alternative way to detect nuclei and their boundaries is training two binary classifiers to detect $inside$ and the boundary separately and then merge the detections together. We apply the same method with our NB model to train the nucleus model and boundary model except that the three-class classification is replaced by binary classification. 
Fig. \ref{fig:three_two} depicts why the NB model outperforms the mixed nucleus model + boundary model. The NB model is able to learn the latent relationships between $inside$, $boundary$ and $background$. That is, there should be no gaps between $inside$ and $boundary$ class and $inside$ should not cross the $boundary$ class. From the samples shown in Fig. \ref{fig:three_two}, we can easily find out that NB model predicts the $inside$ class and $boundary$ class more precisely.

\section{Conclusion}
In this paper, we have presented a state-of-the-art supervised fully convolutional neural network method for nuclei segmentation in histopathological images. First, the histopathological images are normalized into the same color space. To handle the extra-large image issue, one whole image is split into overlapping patches for succeeding processing. Next, we propose a novel nucleus-boundary model to detect nuclei and boundaries on each patch. Then the predictions of all the patches are seamlessly reassembled to build the raw prediction result of the whole image. At the end, we apply a fast and non-parameter post-processing to generate the final nuclei segmentation results. The nucleus-boundary model is trained on very limited number of images and has been tested on the images that may have different appearances. Comparison with the state-of-the-art algorithm shows that our proposed method is accurate, robust, and fast. It is also found that our idea of simultaneous nucleus-boundary identification model can be applied to other image segmentation tasks such as cell segmentation, bacteria segmentation and so on.

\section{Acknowledgement}
We gratefully acknowledge the support of NVIDIA Corporation with the donation of the Titan X Pascal GPU used for this research.


\bibliographystyle{IEEEtran}
\bibliography{dense_unet,digitalpathology}

\begin{thebibliography}{10}
\providecommand{\url}[1]{#1}
\csname url@samestyle\endcsname
\providecommand{\newblock}{\relax}
\providecommand{\bibinfo}[2]{#2}
\providecommand{\BIBentrySTDinterwordspacing}{\spaceskip=0pt\relax}
\providecommand{\BIBentryALTinterwordstretchfactor}{4}
\providecommand{\BIBentryALTinterwordspacing}{\spaceskip=\fontdimen2\font plus
\BIBentryALTinterwordstretchfactor\fontdimen3\font minus
  \fontdimen4\font\relax}
\providecommand{\BIBforeignlanguage}[2]{{%
\expandafter\ifx\csname l@#1\endcsname\relax
\typeout{** WARNING: IEEEtran.bst: No hyphenation pattern has been}%
\typeout{** loaded for the language `#1'. Using the pattern for}%
\typeout{** the default language instead.}%
\else
\language=\csname l@#1\endcsname
\fi
#2}}
\providecommand{\BIBdecl}{\relax}
\BIBdecl

\bibitem{pmid24759275}
M.~Veta, J.~P. Pluim, P.~J. van Diest, and M.~A. Viergever, ``{{B}reast cancer
  histopathology image analysis: a review},'' \emph{IEEE Trans Biomed Eng},
  vol.~61, no.~5, pp. 1400--1411, May 2014.

\bibitem{pmid28066683}
Z.~Gandomkar, P.~C. Brennan, and C.~Mello-Thoms, ``{{C}omputer-based image
  analysis in breast pathology},'' \emph{J Pathol Inform}, vol.~7, p.~43, 2016.

\bibitem{nawaz2015computational}
S.~Nawaz and Y.~Yuan, ``Computational pathology: Exploring the spatial
  dimension of tumor ecology,'' \emph{Cancer letters}, 2015.

\bibitem{chen2015new}
J.-M. Chen, A.-P. Qu, L.-W. Wang, J.-P. Yuan, F.~Yang, Q.-M. Xiang, N.~Maskey,
  G.-F. Yang, J.~Liu, and Y.~Li, ``New breast cancer prognostic factors
  identified by computer-aided image analysis of he stained histopathology
  images,'' \emph{Scientific reports}, vol.~5, 2015.

\bibitem{otsu1975threshold}
N.~Otsu, ``A threshold selection method from gray-level histograms,''
  \emph{Automatica}, vol.~11, no. 285-296, pp. 23--27, 1975.

\bibitem{filipczuk2011automatic}
P.~Filipczuk, M.~Kowal, and A.~Obuchowicz, ``Automatic breast cancer diagnosis
  based on k-means clustering and adaptive thresholding hybrid segmentation,''
  in \emph{Image processing and communications challenges 3}.\hskip 1em plus
  0.5em minus 0.4em\relax Springer, 2011, pp. 295--302.

\bibitem{rother2004grabcut}
C.~Rother, V.~Kolmogorov, and A.~Blake, ``Grabcut: Interactive foreground
  extraction using iterated graph cuts,'' in \emph{ACM transactions on graphics
  (TOG)}, vol.~23, no.~3.\hskip 1em plus 0.5em minus 0.4em\relax ACM, 2004, pp.
  309--314.

\bibitem{al2010improved}
Y.~Al-Kofahi, W.~Lassoued, W.~Lee, and B.~Roysam, ``Improved automatic
  detection and segmentation of cell nuclei in histopathology images,''
  \emph{IEEE Transactions on Biomedical Engineering}, vol.~57, no.~4, pp.
  841--852, 2010.

\bibitem{veta2013automatic}
M.~Veta, P.~J. van Diest, R.~Kornegoor, A.~Huisman, M.~A. Viergever, and J.~P.
  Pluim, ``Automatic nuclei segmentation in h\&e stained breast cancer
  histopathology images,'' \emph{PloS one}, vol.~8, no.~7, p. e70221, 2013.

\bibitem{mouelhi2013automatic}
A.~Mouelhi, M.~Sayadi, F.~Fnaiech, K.~Mrad, and K.~B. Romdhane, ``Automatic
  image segmentation of nuclear stained breast tissue sections using color
  active contour model and an improved watershed method,'' \emph{Biomedical
  Signal Processing and Control}, vol.~8, no.~5, pp. 421--436, 2013.

\bibitem{xu2016stacked}
J.~Xu, L.~Xiang, Q.~Liu, H.~Gilmore, J.~Wu, J.~Tang, and A.~Madabhushi,
  ``Stacked sparse autoencoder (ssae) for nuclei detection on breast cancer
  histopathology images,'' \emph{IEEE transactions on medical imaging},
  vol.~35, no.~1, pp. 119--130, 2016.

\bibitem{sirinukunwattana2016locality}
K.~Sirinukunwattana, S.~E.~A. Raza, Y.-W. Tsang, D.~R. Snead, I.~A. Cree, and
  N.~M. Rajpoot, ``Locality sensitive deep learning for detection and
  classification of nuclei in routine colon cancer histology images,''
  \emph{IEEE transactions on medical imaging}, vol.~35, no.~5, pp. 1196--1206,
  2016.

\bibitem{liao2016automatic}
M.~Liao, Y.-q. Zhao, X.-h. Li, P.-s. Dai, X.-w. Xu, J.-k. Zhang, and B.-j. Zou,
  ``Automatic segmentation for cell images based on bottleneck detection and
  ellipse fitting,'' \emph{Neurocomputing}, vol. 173, pp. 615--622, 2016.

\bibitem{su2014automatic}
H.~Su, F.~Xing, J.~D. Lee, C.~A. Peterson, and L.~Yang, ``Automatic myonuclear
  detection in isolated single muscle fibers using robust ellipse fitting and
  sparse representation,'' \emph{IEEE/ACM Transactions on Computational Biology
  and Bioinformatics}, vol.~11, no.~4, pp. 714--726, 2014.

\bibitem{kharma2007automatic}
N.~Kharma, H.~Moghnieh, J.~Yao, Y.~P. Guo, A.~Abu-Baker, J.~Laganiere,
  G.~Rouleau, and M.~Cheriet, ``Automatic segmentation of cells from
  microscopic imagery using ellipse detection,'' \emph{IET Image Processing},
  vol.~1, no.~1, pp. 39--47, 2007.

\bibitem{qu2014two}
A.~Qu, J.~Chen, L.~Wang, J.~Yuan, F.~Yang, Q.~Xiang, N.~Maskey, G.~Yang,
  J.~Liu, and Y.~Li, ``Two-step segmentation of hematoxylin-eosin stained
  histopathological images for prognosis of breast cancer,'' in
  \emph{Bioinformatics and Biomedicine (BIBM), 2014 IEEE International
  Conference on}.\hskip 1em plus 0.5em minus 0.4em\relax IEEE, 2014, pp.
  218--223.

\bibitem{xing2016automatic}
F.~Xing, Y.~Xie, and L.~Yang, ``An automatic learning-based framework for
  robust nucleus segmentation,'' \emph{IEEE transactions on medical imaging},
  vol.~35, no.~2, pp. 550--566, 2016.

\bibitem{long2015fully}
J.~Long, E.~Shelhamer, and T.~Darrell, ``Fully convolutional networks for
  semantic segmentation,'' in \emph{Proceedings of the IEEE Conference on
  Computer Vision and Pattern Recognition}, 2015, pp. 3431--3440.

\bibitem{ronneberger2015u}
O.~Ronneberger, P.~Fischer, and T.~Brox, ``U-net: Convolutional networks for
  biomedical image segmentation,'' in \emph{International Conference on Medical
  Image Computing and Computer-Assisted Intervention}.\hskip 1em plus 0.5em
  minus 0.4em\relax Springer, 2015, pp. 234--241.

\bibitem{he2016deep}
K.~He, X.~Zhang, S.~Ren, and J.~Sun, ``Deep residual learning for image
  recognition,'' in \emph{Proceedings of the IEEE conference on computer vision
  and pattern recognition}, 2016, pp. 770--778.

\bibitem{naylor2017nuclei}
P.~Naylor, M.~La{\'e}, F.~Reyal, and T.~Walter, ``Nuclei segmentation in
  histopathology images using deep neural networks,'' in \emph{Biomedical
  Imaging (ISBI 2017), 2017 IEEE 14th International Symposium on}.\hskip 1em
  plus 0.5em minus 0.4em\relax IEEE, 2017, pp. 933--936.

\bibitem{kumar2017dataset}
N.~Kumar, R.~Verma, S.~Sharma, S.~Bhargava, A.~Vahadane, and A.~Sethi, ``A
  dataset and a technique for generalized nuclear segmentation for
  computational pathology,'' \emph{IEEE Transactions on Medical Imaging}, 2017.

\bibitem{vahadane2015structure}
A.~Vahadane, T.~Peng, S.~Albarqouni, M.~Baust, K.~Steiger, A.~M. Schlitter,
  A.~Sethi, I.~Esposito, and N.~Navab, ``Structure-preserved color
  normalization for histological images,'' in \emph{Biomedical Imaging (ISBI),
  2015 IEEE 12th International Symposium on}.\hskip 1em plus 0.5em minus
  0.4em\relax IEEE, 2015, pp. 1012--1015.

\bibitem{khan2014nonlinear}
A.~M. Khan, N.~Rajpoot, D.~Treanor, and D.~Magee, ``A nonlinear mapping
  approach to stain normalization in digital histopathology images using
  image-specific color deconvolution,'' \emph{IEEE Transactions on Biomedical
  Engineering}, vol.~61, no.~6, pp. 1729--1738, 2014.

\bibitem{macenko2009method}
M.~Macenko, M.~Niethammer, J.~Marron, D.~Borland, J.~T. Woosley, X.~Guan,
  C.~Schmitt, and N.~E. Thomas, ``A method for normalizing histology slides for
  quantitative analysis,'' in \emph{Biomedical Imaging: From Nano to Macro,
  2009. ISBI'09. IEEE International Symposium on}.\hskip 1em plus 0.5em minus
  0.4em\relax IEEE, 2009, pp. 1107--1110.

\bibitem{cui2016self}
Y.~Cui and J.~Hu, ``Self-adjusting nuclei segmentation (sans) of
  hematoxylin-eosin stained histopathological breast cancer images,'' in
  \emph{Bioinformatics and Biomedicine (BIBM), 2016 IEEE International
  Conference on}.\hskip 1em plus 0.5em minus 0.4em\relax IEEE, 2016, pp.
  956--963.

\bibitem{wang2016automatic}
P.~Wang, X.~Hu, Y.~Li, Q.~Liu, and X.~Zhu, ``Automatic cell nuclei segmentation
  and classification of breast cancer histopathology images,'' \emph{Signal
  Processing}, vol. 122, pp. 1--13, 2016.

\bibitem{glorot2010understanding}
X.~Glorot and Y.~Bengio, ``Understanding the difficulty of training deep
  feedforward neural networks,'' in \emph{Proceedings of the Thirteenth
  International Conference on Artificial Intelligence and Statistics}, 2010,
  pp. 249--256.

\bibitem{pmid29259059}
M.~Herrera de~la Muela, E.~Garcia~Lopez, L.~Frias~Aldeguer, and
  P.~Gomez-Campelo, ``{{P}rotocol for the {B}{R}{E}{C}{A}{R} study: a
  prospective cohort follow-up on the impact of breast reconstruction timing on
  health-related quality of life in women with breast cancer},'' \emph{BMJ
  Open}, vol.~7, no.~12, p. e018108, Dec 2017.

\bibitem{klambauer2017self}
G.~Klambauer, T.~Unterthiner, A.~Mayr, and S.~Hochreiter, ``Self-normalizing
  neural networks,'' \emph{arXiv preprint arXiv:1706.02515}, 2017.

\bibitem{kingma2014adam}
D.~Kingma and J.~Ba, ``Adam: A method for stochastic optimization,''
  \emph{arXiv preprint arXiv:1412.6980}, 2014.

\bibitem{simard2003best}
P.~Y. Simard, D.~Steinkraus, J.~C. Platt \emph{et~al.}, ``Best practices for
  convolutional neural networks applied to visual document analysis.'' in
  \emph{ICDAR}, vol.~3, 2003, pp. 958--962.

\bibitem{paramanandam2016automated}
M.~Paramanandam, M.~O’Byrne, B.~Ghosh, J.~J. Mammen, M.~T. Manipadam,
  R.~Thamburaj, and V.~Pakrashi, ``Automated segmentation of nuclei in breast
  cancer histopathology images,'' \emph{PloS one}, vol.~11, no.~9, p. e0162053,
  2016.

\bibitem{noh2015learning}
H.~Noh, S.~Hong, and B.~Han, ``Learning deconvolution network for semantic
  segmentation,'' in \emph{Proceedings of the IEEE International Conference on
  Computer Vision}, 2015, pp. 1520--1528.

\end{thebibliography}
%

%

\end{document}